\documentclass[10pt,twocolumn,letterpaper]{article}

\usepackage[pagenumbers]{cvpr} 

\usepackage[dvipsnames]{xcolor}

\newcommand{\method}{ShowRoom3D}
\usepackage{afterpage}
\usepackage{caption}
\usepackage{subcaption}
\usepackage{pifont} 
\usepackage{multirow, overpic, textpos, array}
\usepackage{makecell}
\usepackage{color, colortbl}

\usepackage{mathrsfs} 
\usepackage{amsmath} 
\usepackage{amsmath,amssymb,bm}
\usepackage{mathtools}

\newcommand{\E}{\mathbb{E}}

\newcommand{\gL}{\mathcal{L}}
\newcommand{\vepsilon}{\bm{\epsilon}}
\newcommand{\vx}{\mathbf{x}}
\newcommand{\vg}{\mathbf{g}}

\definecolor{cvprblue}{rgb}{0.21,0.49,0.74}
\usepackage[pagebackref,breaklinks,colorlinks,citecolor=cvprblue]{hyperref}

\def\paperID{4914} 
\def\confName{CVPR}
\def\confYear{2024}

\title{\method{}: Text to High-Quality 3D Room Generation Using 3D Priors}

\author{Weijia Mao$^{1}$\quad Yan-Pei Cao$^{2}$\thanks{Corresponding authors}\quad Jia-Wei Liu$^{1}$\quad Zhongcong Xu$^{1}$\quad Mike Zheng Shou$^{1*}$
\\ \vspace{-0.6em} \\ 
$^1$Show Lab, National University of Singapore\quad $^2$ARC Lab, Tencent PCG 
\\ \vspace{-0.6em} \\ 
\url{https://showroom3d.github.io}
}

\begin{document}


\twocolumn[{
\maketitle
\renewcommand\twocolumn[1][]{#1}
\begin{center}
    \centering
    \includegraphics[width=1\textwidth]{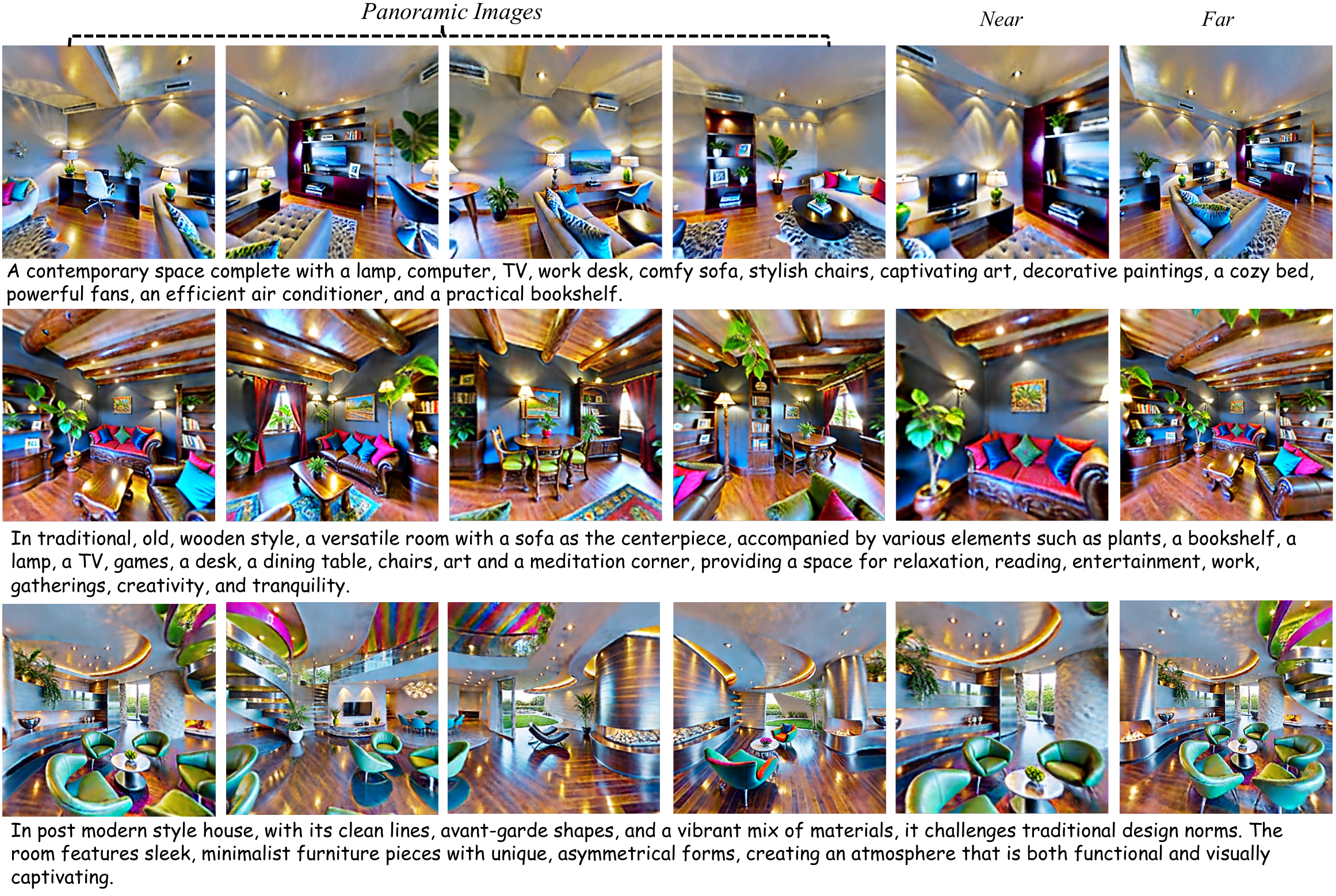}
    \captionof{figure}{
    \textit{\textbf{\method{}:}} A novel method for generating high-quality room-scale scenes that can be rendered at any position.
    }
    \label{fig:teaser}
\end{center}
}]

{
  \renewcommand{\thefootnote}%
    {\fnsymbol{footnote}}
  \footnotetext[1]{Corresponding Author.}
}

\begin{abstract}
We introduce \method{}, a three-stage approach for generating high-quality 3D room-scale scenes from texts. Previous methods using 2D diffusion priors to optimize neural radiance fields for generating room-scale scenes have shown unsatisfactory quality. This is primarily attributed to the limitations of 2D priors lacking 3D awareness and constraints in the training methodology. In this paper, we utilize a 3D diffusion prior, MVDiffusion, to optimize the 3D room-scale scene. Our contributions are in two aspects. Firstly, we propose a progressive view selection process to optimize NeRF. This involves dividing the training process into three stages, gradually expanding the camera sampling scope. Secondly, we propose the pose transformation method in the second stage. It will ensure MVDiffusion provide the accurate view guidance. As a result, \method{} enables the generation of rooms with improved structural integrity, enhanced clarity from any view, reduced content repetition, and higher consistency across different perspectives. Extensive experiments demonstrate that our method, significantly outperforms state-of-the-art approaches by a large margin in terms of user study.
\end{abstract}
    
\section{Introduction}
\begin{figure}[t!]
  \centering
  \includegraphics[width=1\linewidth]{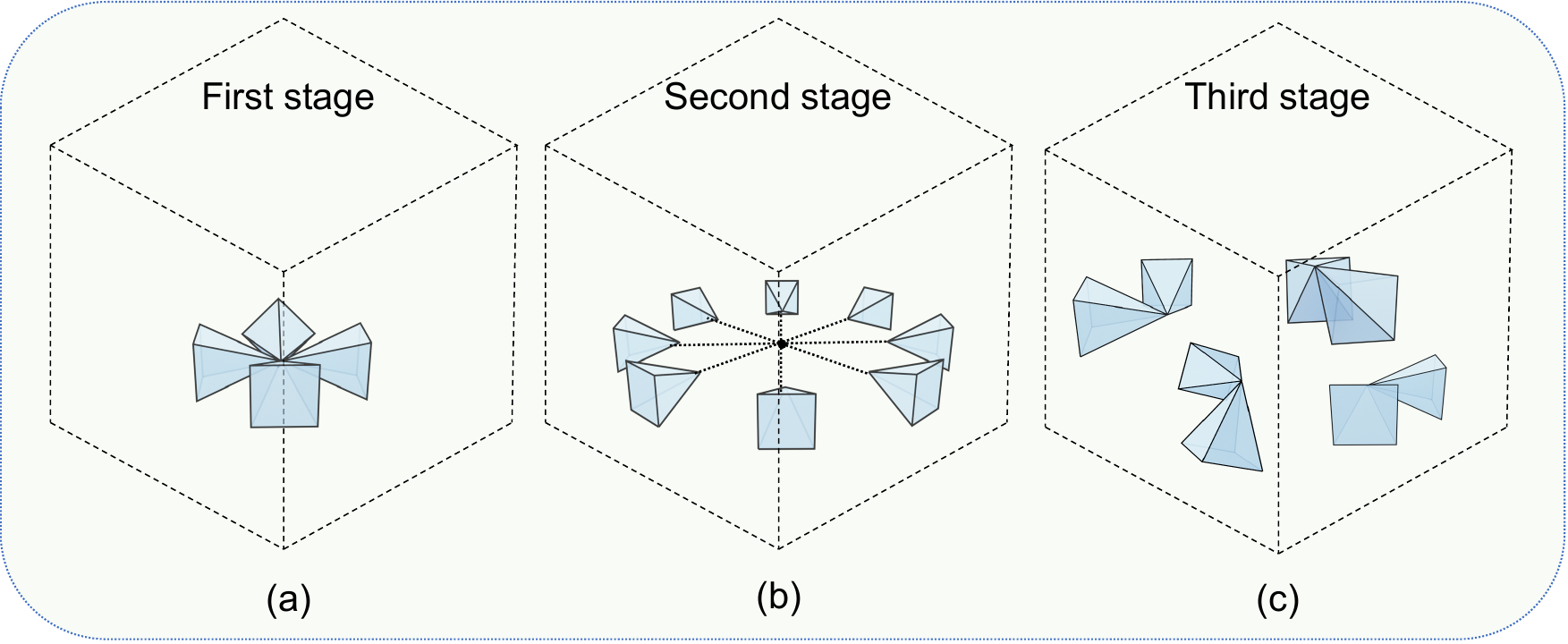}

   \caption{The illustration of every stage's camera sampling method. (a) In the first stage, the camera is positioned at the origin and can rotate freely. (b) In the second stage, the camera is sampled at various positions, but its direction always faces outward from the origin. (c) In the third stage, at different iterations, the camera position and perspective are randomly sampled. Within a single iteration, the cameras remain in the same position.}
   \label{fig:camera_sample}
\end{figure}

The generation of 3D room-scale scenes is crucial for various industries, including VR/AR and Metaverse. In the 2D domain, there has been significant progress in image generation conditioned on user input, thanks to models like Stable Diffusion~\cite{StableDiffusion} and Imagen~\cite{Imagen}. 2D image generation models allow users to control content using prompts and other modalities, such as layouts or poses. However, in the 3D domain, the lack of large-scale 3D datasets has led to methods~\cite{dreamfusion,Magic3D,SJC,HIFA,Prolificdreamer} that combine the text-to-image diffusion model with 3D representations such as NeRF~\cite{NeRF} or DMTet~\cite{DMtet}, often focusing on individual 3D objects generation or their combinations.

However, few of these methods tackle the challenges of 3D room-scale scene generation, as the output needs to be dense, coherent, and encompass all required structures in the views. When applied to 3D room generation, they encounter serious issues such as the Janus problem, unreasonable room structure, style inconsistencies, and more. At the same time, another line of research~\cite{Text2room,scenescape} involves employing the Stable Diffusion inpainting model to generate 3D scenes. However, the consistency within indoor environments is notably subpar, characterized by visible distortions and blurring. 

Recently, MVDiffusion~\cite{MVDiffusion} is proposed to generate the panoramic images of a 3D scene, offering several advantages. (1) It is the first model finetuned on the Stable Diffusion model using the Matterport3D indoor scene dataset~\cite{matterport3d}. It will provide the model with enhanced prior knowledge about the structures and layouts of 3D rooms. (2) MVDiffusion introduces the Correspondence-Aware Attention (CAA) module to ensure consistency between views. This endows the model with 3D awareness, considering the generated images as integral parts of the entire room. However, MVDiffusion is specifically designed to generate only panoramic images of a 3D scene and cannot be used to create a fully realized 3D space. The geometry and structure of the generated scene are not guaranteed.

In our approach, we leverage MVDiffusion in conjunction with NeRF to create a 3D room-scale scene. However, optimizing a NeRF model, which accurately represents the room with high-quality geometry and appearance, using MVDiffusion is not a trivial task. There are two challenges: (1) Confirming the room's geometry and structure while also aiming for rendering from any view during the same training stage tends to yield suboptimal results. (2) The pretrained MVDiffusion model can not effectively handle scenarios where the camera is placed at any position within the room, except for the origin. In such cases, the MVDiffusion panoramic model assumes the camera is at the center of the room, providing inaccurate view guidance for NeRF training.

To address the first challenge, we adopt a progressive view selection approach and divide our training process into three distinct stages. Progressive view selection involves gradually expanding the camera sampling scope in different stages. We first ensure that the geometry of the room is well generated within a limited set of training views, and then we consider rendering the room at any position. As illustrated in Figure~\ref{fig:camera_sample}(a), during the first stage, we position the camera at the center of the room to generate a panoramic view. This initial step is crucial for determining the structure and geometry of the room. In the second stage, we continue to distill the NeRF model, ensuring that the camera consistently faces outward from the origin at any position, as depicted in Figure~\ref{fig:camera_sample}(b). This stage further improves the room's geometry and enables rendering from multiple viewpoints. In the third stage, we sample the cameras at any position and apply rotations to refine the NeRF model at different iterations. During each iteration, we randomly sample cameras from the same position, as shown in Figure~\ref{fig:camera_sample}(c). Ultimately, this process will yield a NeRF model capable of rendering the generated rooms from any position and at any rotation.

To address the second challenge, we introduce a pose transformation method to address situations where the sampled cameras are not at the origin. This transformation will provide an equivalent camera pose to the MVDiffusion model, as opposed to using the real pose. The equivalent camera will share the similar view with the real one. This method ensures that MVDiffusion provides accurate view guidance, even when the sampled cameras' positions are not at the origin.

In summary, our key contributions are as follows: (1) We are the first to explore the utilization of 3D diffusion prior for generating high-quality 3D room-scale scenes using the SDS method. (2) We present a three-stage training pipeline, incorporating distinct camera sampling methods in each stage and pose transformation in the second stage to improve the clarity and aesthetics of the generated room. (3) Our method enables the generation of state-of-the-art 3D room-scale scenes, showcasing not only more compelling geometry and appearance but also a more reasonable room structure and reduced content repetition. Furthermore, it exhibits the capability to render views across a larger space, surpassing the capabilities of previous methods.

\section{Related Work}
\textbf{3D Content Generation.}
The emergence of NeRF~\cite{NeRF} has significantly improved the quality of novel view synthesis in the 3D domain. NeRF-based models~\cite{NeRF,DNeRF,Nerfies,DeVRF,DVGO,liu2023hosnerf,weng2022humannerf} integrate volume rendering algorithms with MLPs or voxels to predict color and opacity. In the field of 3D generation, many works~\cite{jain2022zero,EG3D,GRAF,GRAM,giraffe,PV3D} have combined 2D unconditional generative models with NeRF to create 3D contents. 

DreamFusion~\cite{dreamfusion} proposes score distillation sampling (SDS) to utilize a 2D text-to-image model to optimize the NeRF~\cite{NeRF} model for generating 3D objects. However, the generated content is often oversaturated and plagued by the Janus problem (multi-head problem). Subsequent research~\cite{Prolificdreamer,HIFA,Perp-Neg}endeavors aimed to improve the quality of 3D content and alleviate oversaturation and the Janus problem. Prolificdreamer~\cite{Prolificdreamer} introduced an improved method, VSD, which utilizes LoRA~\cite{LoRA} to train with the NeRF model, effectively alleviating oversaturation. Set-the-scene~\cite{set-the-scene} and CompoNeRF~\cite{CompoNeRF} utilize the SDS distillation method to create scenes with straightforward object compositions. However, these approaches are limited to generating very basic scenes comprising only a few objects. Several studies~\cite{vox-e,dreameditor,SKED,liu2023dynvideo} also employ the SDS-based approaches to edit the NeRF conditioned on users' prompts. 

Several work has developed 3D generation models based on 2D generation models using 3D datasets. Zero123~\cite{zero123} uses the Objvarse~\cite{objvarse} dataset to finetune the Stable Diffusion model~\cite{StableDiffusion}, enabling it to generate view-consistent images. Magic123~\cite{Magic123} leverages both the Stable Diffusion model and the Zero123 model as 2D and 3D priors to optimize NeRF, resulting in a consistent structure for 3D objects. MVDream~\cite{MVDream} redesigns the Stable Diffusion architecture, incorporating a 3D-aware attention module. It uses the Objvarse dataset~\cite{objvarse} for finetuning to generate high-quality content that mitigates the Janus problem. However, these efforts predominantly center around 3D object generation, with limited focus on generating high-quality indoor scenes.

\noindent\textbf{Indoor Scene Generation.} Historically, several works~\cite{Look-outside-the-room,pose-guided-diffusion-models,synsin,geogpt,ses3d} have utilized real indoor scene datasets like Matterport3D~\cite{matterport3d}, ScanNet~\cite{Scannet}, and RealEstate~\cite{RealEstate} datasets to train generative models, such as GANs or autoregressive transformers, for synthesizing novel views. However, these approaches have primarily focused on generating novel views within indoor scenes, without capturing the entire 3D indoor environment. As a result, the generated content quality has been suboptimal in consistency. At the same time, Scenescape~\cite{scenescape}, Text2room~\cite{Text2room} and Text2NeRF~\cite{text2nerf} utilize the inpainting function of Stable Diffusion models to generate 3D scenes. Nevertheless, these methods still face challenges in maintaining style and content consistency, as the generation of each image is conditioned solely on the previous image.

MVDiffusion~\cite{MVDiffusion} aims to address this issue by introducing a correspondence attention module to capture relationships between views. The camera poses in the generated views will be fed into MVDiffusion model. Besides, it finetunes the Stable Diffusion model using the Matterport3D~\cite{matterport3d} dataset. However, it only can generate view-consistent panoramic images. There are limitations in assuring the geometry of the room and rendering the room from any view.

\begin{figure*}[t!]
  \centering
  \includegraphics[width=1\linewidth]{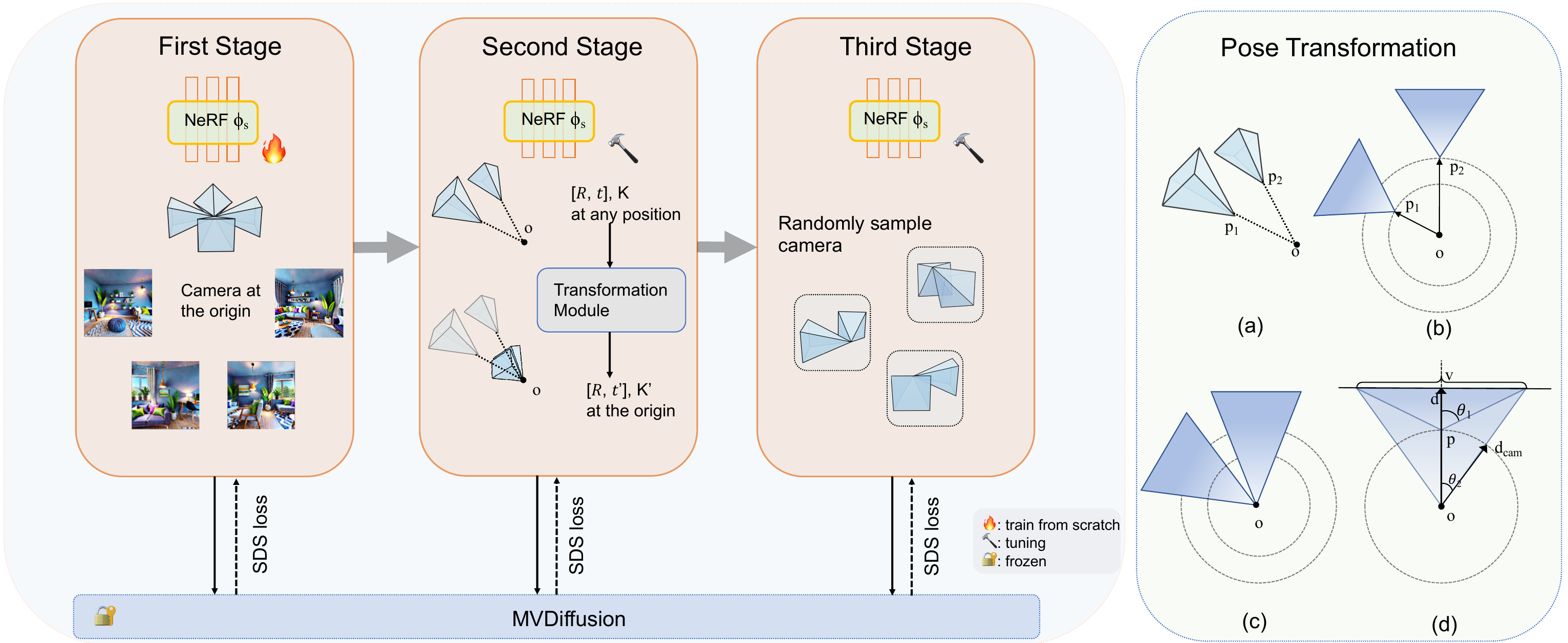}

   \caption{Method overview: \textbf{Left}: Our three-stage training pipeline. First Stage: the camera will be at the center of the room and rotate any degree. Second stage: the camera will be at any position and face outward from the center $\mathrm{o}$. Third stage: the camera will be at any position and rotate any degree.
   \textbf{Right}: It introduces the pose transformation module in the second stage. (a) shows the camera sampling method in the second stage. $\mathrm{p1}$ and $\mathrm{p2}$ are sampled points at one iteration and $\mathrm{o}$ is the center of the room in 3D space. (b) shows the perspective of (a) observed from a 2D plane. (c) shows the two new cameras after the pose transformation. (d) shows the specific scenes observed by the camera at each sampling point. $\mathrm{v}$, $d_\mathrm{cam}$, $\mathrm{d}$ represent the observed view, the distance from the camera position to the origin and the depth of the view. $\theta_1$ and $\theta_2$ represent the FOV of two cameras.  }
   \label{fig:pipeline}
\end{figure*}

\section{Method}

In this section, we first briefly introduce the score distillation sampling (SDS) method and the MVDiffusion model which our method is based on (Section.~\ref{sec:preliminary}). Then we will introduce our task setting and our approach ShowRoom3D that adopts three-stage training procedure (Section.~\ref{sec:our_method}). Our pipeline is shown in Figure~\ref{fig:pipeline}.

\subsection{Preliminary}\label{sec:preliminary}
\textbf{Text-to-3D Generation by Score Distillation Sampling (SDS)~\cite{dreamfusion}.}
SDS is a promising method that combines 2D generative models, such as Stable Diffusion~\cite{StableDiffusion}, with 3D representations, like Neural Radiance Fields (NeRF)~\cite{NeRF}, to generate 3D objects and scenes solely based on textual prompts. When using a pretrained text-to-image model, noise is introduced into the NeRF's output. Then the noisy image $\vx_t$, a time step $t$, a text embedding $y$ are then fed into a U-Net architecture to predict the noise, denoted as $\hat{\vepsilon}$. The optimization of NeRF's parameters, denoted as $\phi$, is based on minimizing the loss between the noise and the predicted noise. In recent times, distillation methods have been effectively applied in the field of 3D generation, producing significant improvements.
Its gradient is approximated by
\begin{equation}
\label{eq:sds}
    \nabla_{\phi} \gL_{\mbox{\tiny SDS}}(\phi) \approx \E_{t,\vepsilon,c}\,\left[\omega(t)(\hat{\vepsilon}(\vx_t,t,y) - \vepsilon)\frac{\partial\vg(\phi,c)}{\partial\phi}\right]\mbox{,} 
\end{equation}
where $\omega(t)$ is a weighting function. Despite SDS has made significant strides in 3D object generation, how to use the SDS method specifically for room-scale scene generation remains underexplored.

\noindent\textbf{MVDiffusion~\cite{MVDiffusion}.}
The MVDiffusion model is a multi-view consistency generation model built upon Stable Diffusion. One of the crucial components that ensures view consistency is the correspondence-aware attention mechanism (CAA module), which is employed to capture the relationship between adjacent views. This mechanism takes into account both the source feature maps, denoted as $F$, and the target feature maps, denoted as $F^l$.

For each source feature, MVDiffusion leverages the relative camera pose to determine the location of the corresponding target feature. $s$ and $t$ represent the source and target pixel. $\bar{F}(\mathbf{s})$ is the source feature with the positional encoding. $l$ is the number of target feature maps and $\bar{F}^l(t^{l}_{*})$ is the target feature with the positional encoding. $\mathcal{N}(\mathbf{t}^l)$ is the neighborhood of the target pixel. Subsequently, the output of the attention mechanism is calculated as follows,

\begin{align}
\begin{split}
\mathbf{Q} &= \mathbf{W_Q}\bar{F}(\mathbf{s}), \\
\mathbf{K} &= \mathbf{W_K}\bar{F}^l(t^{l}_{*}), \\
\mathbf{m} &= \sum_l\sum_{t^{l}_{*} \in \mathcal{N}(\mathbf{t}^l)} \mbox{SoftMax}(\mathbf{Q} \cdot \mathbf{K}) \cdot \mathbf{W_V}\bar{F}^l(t^{l}_{*}),
\end{split}
\end{align}

where $\mathbf{W_Q}$, $\mathbf{W_V}$, and $\mathbf{W_K}$ represent the query, value, and key components of the attention mechanism. For clarity, we do not introduce some specific details, such as position encoding and adding integer displacements. When we use the MVDiffusion model to generate the panoramic images, the prompt and the camera poses of images need to be fed into it. The camera poses will be used in the CAA module to calculate the relationship between the source feature and the target features. 

\subsection{ShowRoom3D}\label{sec:our_method}

\noindent\textbf{Task Setting.} In this paper, we propose a new method to generate the high-quality room-scale scene that can be rendered at any position. Given by text prompts $y$, our objective is to generate a 3D room represented by a NeRF model $\Phi$ using the pretrained MVDiffusion model $M$. We choose Instant-NGP~\cite{instant-ngp} as our NeRF representation due to its numerous advantages, including rapid coverage speed and the ability to reconstruct complex geometries.

\noindent\textbf{First Stage.} 
As the MVDiffusion model initially generates a panoramic scene, we employ the SDS method to optimize a NeRF model, specifically designed to represent a room-scale scene panorama. Initially, we position the camera at the world coordinate center and randomly sample the rotation degree to obtain the camera pose like in Figure~\ref{fig:camera_sample}(a). This camera pose is then fed into both the NeRF model and the correspondence-aware attention module of the MVDiffusion model. So the gradient can be approximated by 
\begin{equation}
\begin{aligned}
\label{eq:sds}
    \nabla_{\phi} \gL(\phi) \approx \E_{t,\vepsilon,c}\,\left[\omega(t)(\hat{\vepsilon}(\vx_t,t,y,c) - \vepsilon)\frac{\partial\vg(\phi,c)}{\partial\phi}\right]\mbox{.} 
\end{aligned}
\end{equation}

In this equation, $y$ represents the prompt, and $c$ represents the camera pose, including the camera's extrinsics and intrinsics. We will input $c$ into the NeRF model to obtain the rendered image. Subsequently, we introduce noise to the rendered image, resulting in the noisy latent $\vx_t$. This noisy latent $\vx_t$, the prompt $y$, the camera pose $c$ and the timestep $t$ are then fed into the pretrained MVDiffusion model's U-Net module and CAA module for noise prediction and gradient calculation.

\noindent\textbf{Second Stage.} After the first stage, we obtain the panoramic NeRF, which initially determines the geometry and layout of the room. We can rotate any degree to render this room, however, it restricts the camera's position, preventing us from rendering the room from any position. At the same time, the geometry of the room is also subpar due to the limited training views.

In the second stage, our goal has two aspects. Firstly, this stage serves to improve the geometry and room layout by adding training views.  Secondly, we can render the room in a larger space compared to the first stage. We modify the camera sampling method, as shown in the Figure~\ref{fig:camera_sample}(b). Now, the camera can be sampled at any position, and it always faces outward from the origin. 

The challenge lies in ensuring that MVDiffusion offers precise view guidance even when cameras are not positioned identically. To solve this problem, we propose pose transformation to obtain a new camera pose to be fed into the MVDiffusion model, rather than using the real camera pose directly. For any sampled camera not positioned at the center, we will employ pose transformation to obtain an equivalent new pose at the center. The core idea is that the new camera pose at the center will have a smaller field of view (FOV). Even if the new camera is farther from the scene, the smaller FOV will help it see the similar view compared with the old camera. Next, we will demonstrate the procedure for calculating the new camera pose.

In Figure~\ref{fig:pipeline}(a)(b), $p1$ and $p2$ are two points we sampled at one iteration, which represent the camera positions. In Figure~\ref{fig:pipeline}(d), let point $o$ be the center of the room and the world coordinate, and point $p$ represent the camera position. $\theta_1$ denotes the FOV of the camera at position $p$. $v$ represents the region visible to the camera and $d$ is the averaged depth of this region. The real camera is located at $p$. Pose transformation is used to calculate the new camera at the position $o$. $\theta_2$ is the FOV of the new camera at position $o$. To ensure that two cameras have a similar view $v$, we calculate $\theta_2$ using the following formulas.
\begin{align}
\begin{split}
\frac{v}{2} &= \tan\theta_2 \cdot d\mbox{,}\\
\frac{v}{2} &= \tan\theta_1 \cdot (d - d_\mathrm{cam})\mbox{,}\\
\theta_2 &= \arctan\left(\frac{\tan\theta_1 \cdot (d - d_\mathrm{cam})}{d}\right)\mbox{,}\\
\end{split}
\end{align}
where $d_\mathrm{cam}$, $\theta_1$ are known parameters and the depth $d$ can be obtained by the first stage's prior. So we can calculate the new $\theta_2$. Then we can use the rendered view from the camera at position $o$ approximate the view from the camera at position $p$ by changing the FOV $\theta_1$ to $\theta_2$. 

After the pose transformation, we obtain two different yet similar camera poses $c(R,t,K)$, which represents the real camera pose fed into the NeRF model, and $c'(R,t',K')$, an approximated new camera pose fed into MVDiffusion. The rotation matrix $R$ remains constant, while the camera position $t$ and the intrinsics $K$ undergo changes. Then we can confirm that even if the sampled points are at different positions $p1$ and $p2$, the new cameras' positions fed into MVDiffusion are the same position $o$. This ensures that MVDiffusion provides the most accurate guidance about the relationships between views.

\noindent\textbf{Third Stage.}
After the two-stage training process, we have established the geometry and structure of the room, enabling rendering from a wide range of perspectives. However, some issues remain after the two-stage training. The first concern is the approximated nature of the pose transformation method, which may not yield highly accurate results. The depth prior is not accurate and the rendering views from the real camera pose and the approximated new camera pose are not perfectly identical. Furthermore, during the first two stages of training, the camera consistently faces outward from the origin, resulting in some missing training views. 

In the third stage, we position the camera freely and applied various degrees of rotation to further finetune the NeRF model. Following this stage, the NeRF becomes a versatile renderer capable of handling scenes from any position and at any rotation. To be specific, as shown in the Figure~\ref{fig:camera_sample}(c), we sample two points at the same position at one iteration fed into NeRF and MVDiffusion. It will ensure MVDiffusion provide the relatively accurate guidance when the sampled cameras are the same position, even if they are not at the center. At different iterations, the camera position and perspective are randomly sampled.

\begin{figure*}[t!]
  \centering
  \includegraphics[width=1\linewidth]{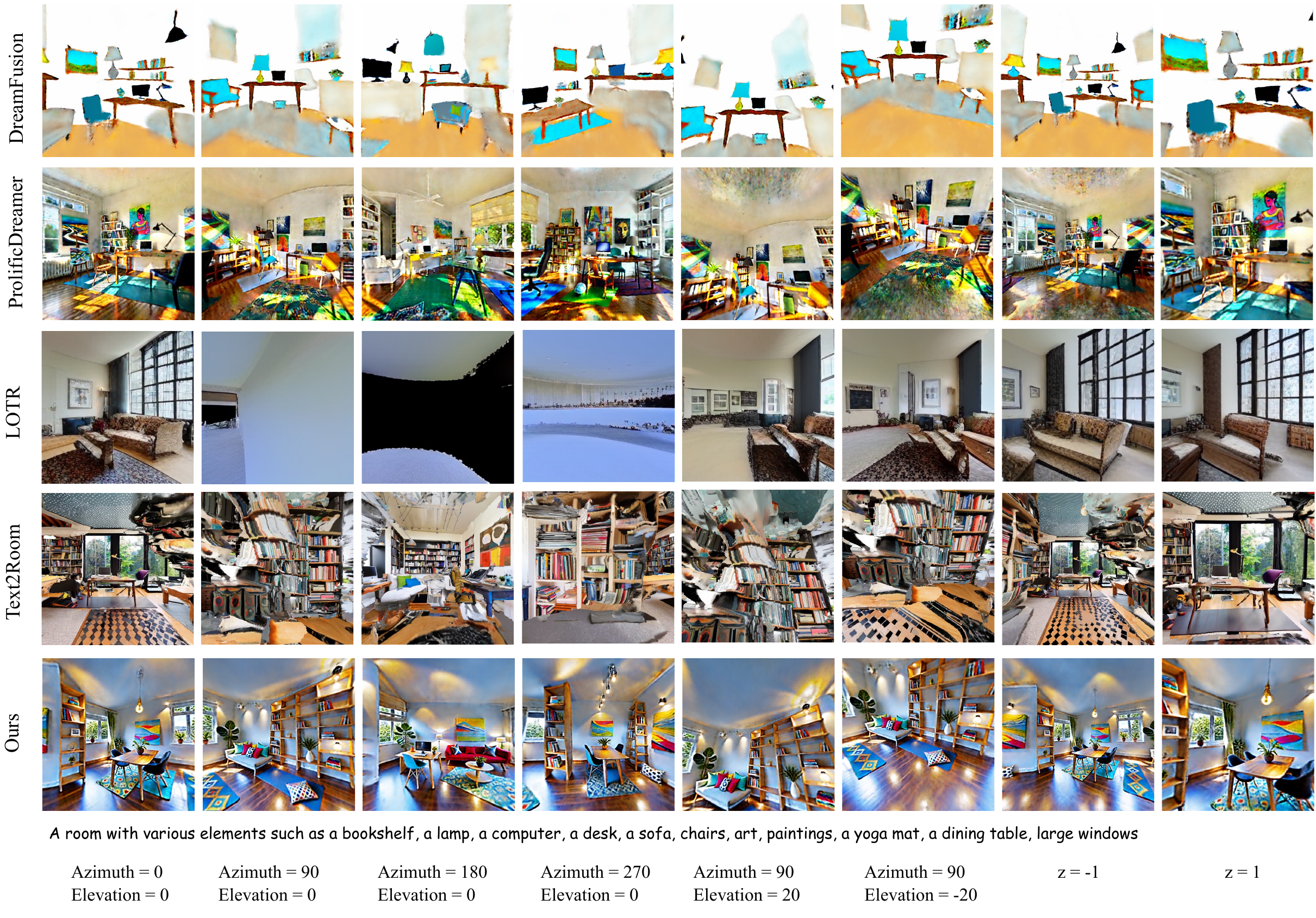}
   \caption{Qualitative comparisons of ShowRoom3D and state-of-the-art approaches.}
   \label{fig:baseline_comparison}
\end{figure*}

\section{Experiments}
\subsection{Implemention details}
 In our approach, we utilize the panoramic model from MVDiffusion~\cite{MVDiffusion} as the foundational model for optimizing the NeRF~\cite{NeRF}. We maintain the same world coordinate system and camera coordinate system as MVDiffusion. We implement the SDS method in the threestudio framework~\cite{threestudio2023}. The first stage of NeRF training requires 10,000 iterations, the second stage needs 15,000 iterations and the third stage needs 5000 iterations.
We utilize a single NVIDIA RTX 3090 GPU to train the three-stage NeRF model.
To mitigate the issue of oversaturation introduced by the SDS method, we draw inspiration from MVDream~\cite{MVDream} and employ similar techniques. Specifically, we anneal the timestep to control the noise added to the NeRF's output. Another strategy involves using negative prompts to guide the training process. More details will be provided in the supplementary materials.

\subsection{Baselines}
We select four state-of-the-art works to compare with our method. (1) \textit{DreamFusion}~\cite{dreamfusion} is the first work to use the text-to-image diffusion model to optimize a 3D object. (2) \textit{ProlificDreamer}~\cite{Prolificdreamer} introduces a new method, VSD, for optimizing a NeRF. VSD combines the vanilla Stable Diffusion model with the LORA~\cite{LoRA} model to jointly optimize a NeRF, resulting in the alleviation of oversaturation phenomena. (3) \textit{Text2Room}~\cite{Text2room} is another method that leverages a 2D diffusion model to generate 3D scenes. It utilizes the Stable Diffusion inpainting model to generate 2D images sequentially. (4) \textit{Look Outside the Room}~\cite{Look-outside-the-room} is a method that generates novel view images based on previously generated images and camera poses. It trains an autoregressive transformer model from scratch using the Matterport3D~\cite{matterport3d} dataset and RealEstate~\cite{RealEstate} dataset.
\begin{figure*}[t]
  \centering
  \includegraphics[width=1\linewidth]{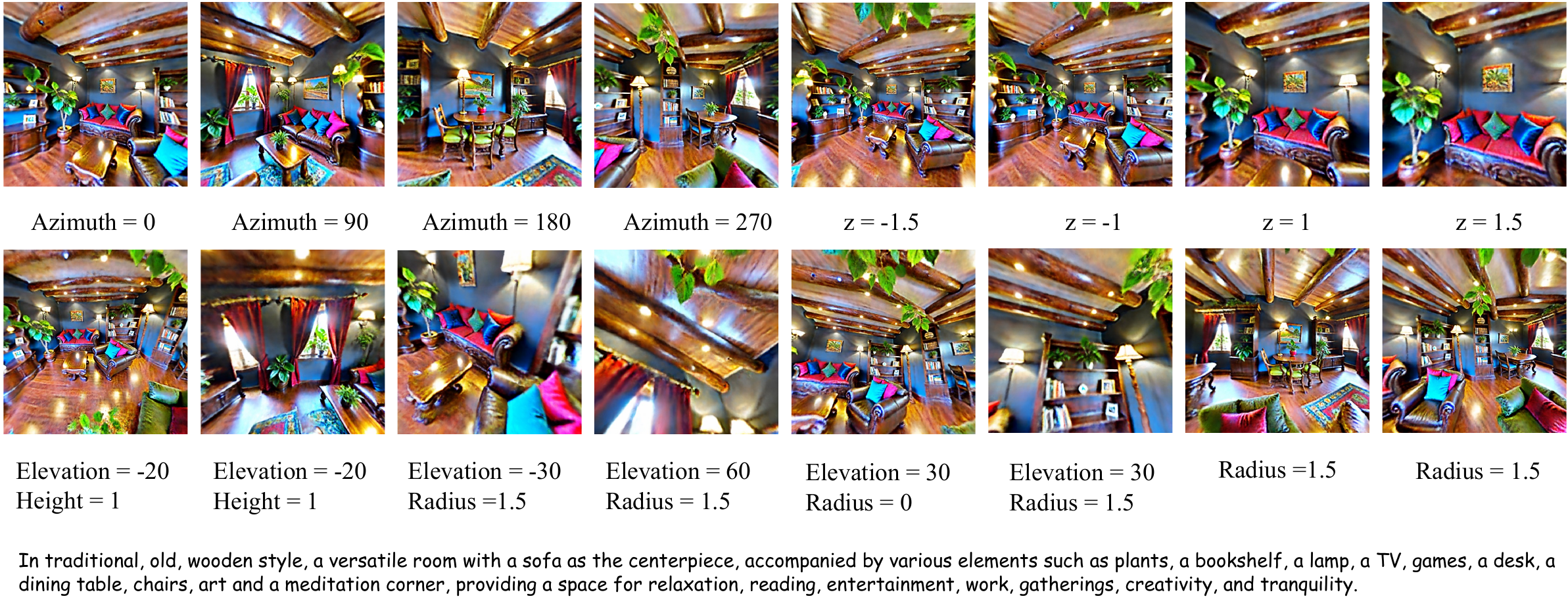}
   \caption{More views of our result. }
   \label{fig:our_result}
\end{figure*}

\newcommand{\tablefirst}[0]{\cellcolor{myred}}

\begin{table*}

\vspace{-1mm}
\centering

\centering
\setlength{\tabcolsep}{2.5pt}

\begin{tabular}{l||cc||ccc}

\toprule
& \multicolumn{ 2 }{c||}{
  \makecell{
  \text{\small Metrics }
  }
}
& \multicolumn{ 3 }{c}{
  \makecell{
  \text{\small Human Preference}
  }
}

\\

& \multicolumn{1}{c}{ \footnotesize CLIP Score ($\uparrow$) }
& \multicolumn{1}{c||}{ \footnotesize Aesthetic Score ($\uparrow$) }
& \multicolumn{1}{c}{ \footnotesize Textual Alignment ($\uparrow$) }
& \multicolumn{1}{c}{ \footnotesize Consistency ($\uparrow$) }
& \multicolumn{1}{c}{ \footnotesize Overall Quality ($\uparrow$) }

\\
\hline

  DreamFusion~\cite{dreamfusion}
  &$23.56$
  &$4.65$
  
  &$2.75$
  &$2.80$
  &$2.55$
  
  \\ 
  ProlificDreamer~\cite{Prolificdreamer}
  &$22.45$
  &$4.98$
  
  &$3.10$
  &$3.12$
  &$2.87$
  
  \\ 
  Text2Room~\cite{Text2room}
  &$20.41$
  &$5.21$
  
  &$3.34$
  &$2.97$
  &$3.20$

  \\ 
  LOTR~\cite{Look-outside-the-room}
  &$13.12$
  &$4.23$
  
  &$1.25$
  &$1.21$
  &$1.78$

  \\ 
  \method{} (Ours)
  &$\mathbf{25.62}$
  &$\mathbf{5.56}$
  
  &$\mathbf{4.55}$
  &$\mathbf{4.89}$
  &$\mathbf{4.59}$

  \\ \bottomrule

\end{tabular}

\caption{
Quantitative comparisons of \method{} against state-of-the-art approaches.
\label{tab:sota_comparison}
}


\vspace{-2mm}
\end{table*}

\subsection{Qualitative Results}
In Figure~\ref{fig:baseline_comparison}, we present RGB rendering images of scenes with the crucial coordinate information for our method and baselines. The initial four images depict a panoramic view of the entire room, followed by four additional perspectives in the subsequent set of images. DreamFusion and ProlificDreamer struggle to generate the correct room structure, exhibiting issues such as the Janus problem and style inconsistency. This results in blurriness in certain views. Text2Room results in view inconsistencies, stretching, and blurring due to its inpainting method. Meanwhile, LOTR fails to create panoramic scenes and exhibits poor content consistency between frames. Additionally, we present additional views of the rooms generated by our results in Figure~\ref{fig:our_result}. Displaying 16 views of a room, it demonstrates that our method produces not only high-quality rooms but can also render in a larger space. We have briefly annotated the crucial coordinate information, and further results comparing baselines with our method can be found in the supplementary material.

\subsection{Quantitative Results}
We compute the CLIP Score~\cite{clip} and aesthetic score~\cite{aesthetic} for 120 RGB renderings of each scene and conduct a user study for scene evaluation. The CLIP score is employed to assess the alignment between rendered views and provided prompts. Additionally, the aesthetic score introduced by LAION measures the aesthetic quality of the generated images. Recent methods~\cite{sdxl} have demonstrated its authenticity, surpassing the reliability of FID. We present quantitative results, averaged across multiple scenes, in Table~\ref{tab:sota_comparison}. In this table, we observe that our method achieves the highest averaged CLIP score which means our method can generate scenes that closely align with user prompts and the highest averaged aesthetic score which means our results maintain the highest quality. 

Additionally, in our user study, we leverage Amazon MTurk~\footnote{https://requester.mturk.com/}to recruit 24 participants to rank the results obtained from our methods and other baselines. The rendered views from the generated 3D scene are evaluated across three aspects: overall quality, text alignment, and style consistency. Scores are calculated with a rating of 5 for the best-ordered view and 1 for the last. In the end, we gather a total of 259 data points to calculate the final scores. As illustrated in Table~\ref{tab:sota_comparison}, we outperform other methods by a substantial margin, indicating our superior performance in terms of overall quality, text alignment, and consistency.

\begin{figure*}[t!]
  \centering
  \includegraphics[width=1\linewidth]{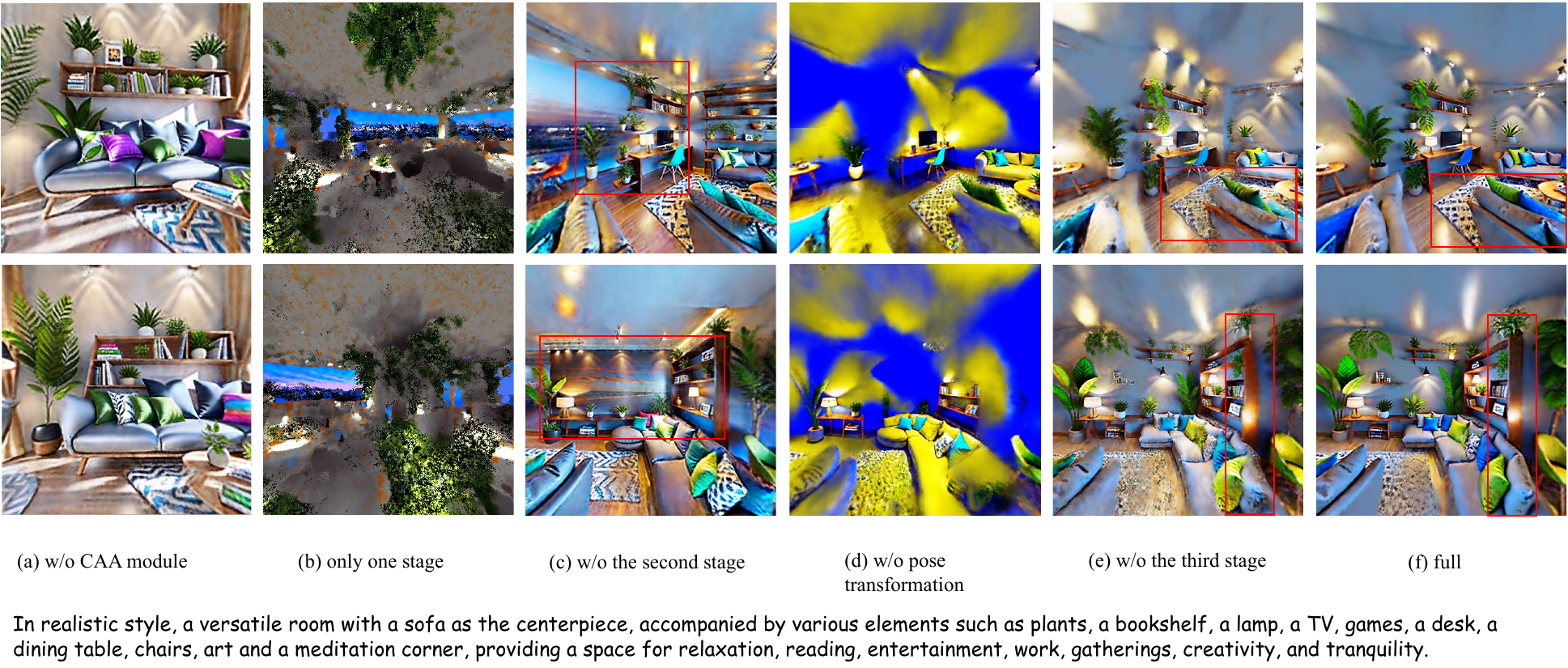}
   \caption{Ablation study on each proposed component. We choose two different views of a room. (a) We use the SD(finetuned on Matterport3D) model without CAA module to follow our three-stage pipeline. The quality of the room is also high but there are many repetitive contents in different views. Results in (b), (c), (d), and (e) demonstrate the impact of removing each corresponding component, showcasing varying degrees of quality degradation. Finally, (f) presents the result of our full method.}
   \label{fig:ablation2}
\end{figure*}

\begin{table}

\vspace{-1mm}
\centering
\resizebox{\linewidth}{!}{

\centering
\setlength{\tabcolsep}{2.5pt}

\begin{tabular}{l||cc}

\toprule
& \multicolumn{ 2 }{c}{
  \makecell{
  \text{\small Metrics }
  }
}

\\

& \multicolumn{1}{c}{ \footnotesize CLIP Score ($\uparrow$) }
& \multicolumn{1}{c}{ \footnotesize Aesthetic Score ($\uparrow$) }

\\
\hline

  Ours w/o CAA module
  &$28.23$
  &$\mathbf{5.62}$

  \\ 
  Ours only one stage
  &$14.48$
  &$4.91$
  
  \\ 
  Ours w/o the second stage
  &$27.94$
  &$5.21$
  \\ 
  Ours w/o pose transformation
  &$27.63$
  &$5.31$
  \\

  Ours w/o the third stage
  &$26.43$
  &$5.44$

  \\ 
  \method{} (Ours full)
  &$\mathbf{28.82}$
  &$5.59$

  \\ \bottomrule

\end{tabular}
}

\caption{
Quantitative ablations of ShowRoom3D.
\label{tab:ablation_comparison}
}


\vspace{-7mm}
\end{table}

\subsection{Ablations}
Our method's key components include three-stage training and pose transformation in the second stage. We also calculate CLIP Score and aesthetic score of every ablation part, as shown in Table~\ref{tab:ablation_comparison}. Our method get the highest CLIP Score and the second highest aesthetic score. Next we will illustrate each component. 

\noindent\textbf{The Effect of CAA Module.}
We just use the Stable Diffusion(finetuned on the Matterport3D dataset) without CAA module as a prior to follow our three-stage pipeline. In Figure~\ref{fig:ablation2}(a), we investigate the impact of the correspondence attention module. It is evident that the use of the correspondence attention module mitigates the Janus problem and improves content diversity of the room in our method, as shown in Figure~\ref{fig:ablation2}(f).

The scores of our method are similar to those without the CAA module, proving the utility of our three-stage training pipeline for both MVDiffusion and Stable Diffusion. While our method's aesthetic score is slightly lower than without the CAA module, this discrepancy may be attributed to the limitations of the two metrics used, as they cannot effectively assess the Janus problem.

\noindent\textbf{The Effect of Training with Only One Stage.}
In Figure~\ref{fig:ablation2}(b), we investigate the impact of three-stage training. It is evident that if we employ the SDS method to optimize the NeRF with only one stage, it struggles to generate meaningful content. This is primarily due to the CAA module's inability to correctly handle the camera pose, resulting in failed training.

\noindent\textbf{The Effect of the Second Stage.}
As depicted in Figure~\ref{fig:ablation2}(c), if we train the NeRF model without the second stage and proceed directly to the third stage after the first stage's training, the geometric quality deteriorates. Some furniture pieces may not be generated effectively, resulting in suboptimal wall and shelf shapes, as illustrated in Figure~\ref{fig:ablation2}(c).

\noindent\textbf{The Effect of Pose Tranformation.}
In the Figure~\ref{fig:ablation2}(d), we explore the consequences of further NeRF optimization without pose transformation. The results indicate that continued NeRF optimization can lead to training instability and sometimes the training will fail because the noise from MVDiffusion model will not provide the accurate view guidance.

\noindent\textbf{The Effect of the Third Stage.}
As depicted in Figure~\ref{fig:ablation2}(e), we investigate the significance of the third stage. Without the third training stage, the quality of rendered views in this room deteriorates because in the first two stages, the camera poses are not randomly sampled, resulting in some missing views that are left uncovered.

\section{Conclusion}
In this work, we introduce \method{}, a three-stage pipeline using a 3D diffusion prior, MVDiffusion, to optimize NeRF for the generation of high-quality 3D room-scale scenes. We employ progressive view selection 
approach in three stages. During the second stage, we utilize pose transformation to ensure accurate guidance from MVDiffusion. As a result, we can produce high-quality room-scale scenes which can be rendered at any position.

\noindent\textbf{Limitations.} Our approach enables the generation of high-quality 3D room-scale scenes from texts. However, there are some limitations. Firstly, similar to previous methods, our approach produces oversaturated results due to the SDS loss, despite employing certain training techniques to alleviate this occurrence. Secondly, our method is time-consuming due to the three-stage training process.

\noindent\textbf{Acknowledgement.} We thank Ziteng Gao, Hai Ci and Yiquan Chen for their helpful discussions.

{
    \small
    \bibliographystyle{ieeenat_fullname}
    \bibliography{main}
}
\clearpage

\maketitlesupplementary

This supplementary mainly includes the implementation details of ShowRoom3D and other baselines, more comparisions of ShowRoom3D against other baselines and more ablation results.

Furthermore, we also provide a \textbf{supplementary video} to show 360° free-viewpoint renderings of the rooms from ShowRoom3D and the comparisons of ShowRoom3D against baselines.

\section*{A. More Implementation Details of ShowRoom3D}
We align our coordinates with MVDiffusion~\cite{MVDiffusion} by adopting the same right-handed world coordinate system and camera coordinate system. Specifically, the x-axis faces forward, the z-axis faces right, and the y-axis faces up. We also take the annealing time tragedy to alleviate the oversaturation phenomenon. In the first stage, the maximum time step is reduced from 0.6 to 0.02, while the minimum timestep is adjusted from 0.98 to 0.7. In the second stage, for the first 10,000 iterations, the maximum and minimum timestep values are set to 0.7 and 0.02, respectively. For the subsequent 5,000 iterations and during the third stage, the maximum and minimum timestep values become 0.4 and 0.02. We also utilize negative prompts to guide the optimization of NeRF. The negative prompt includes descriptors such as `ugly, bad anatomy, blurry, pixelated, obscure, unnatural colors, poor lighting, dull, unclear, cropped, lowres, low quality, artifacts, duplicate, morbid, mutilated, poorly drawn face, deformed, dehydrated, bad proportions.'

\section*{B. More Implementation Details of Baselines}
\noindent\textbf{DreamFusion~\cite{dreamfusion} and ProlificDreamer~\cite{Prolificdreamer}.}  To ensure a fair comparison with our method, we adopt the same right-handed world coordinate system and align the camera coordinate system accordingly. In this setup, the x-axis points forward, the z-axis points right, and the y-axis points up. While DreamFusion initially operates in an object-centered camera system for 3D object generation, we modify it to a camera-centered system for consistency.

\begin{figure}[t!]
  \centering
  \includegraphics[width=0.95\linewidth]{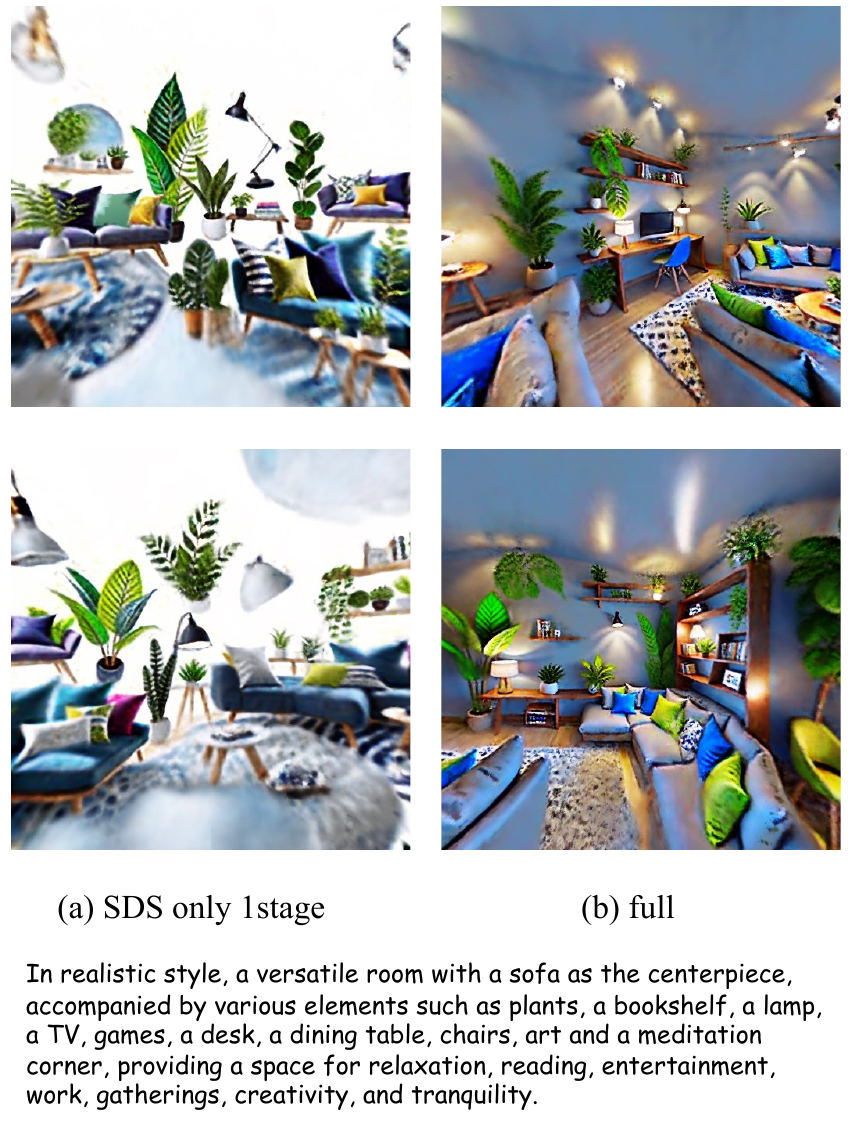}
   \caption{Ablation study on one more component.}
   \label{fig:ablation}
   \vspace{-5mm}
\end{figure}


\noindent\textbf{Text2Room~\cite{Text2room}.} We adhere to the Text2Room pipeline to generate the room mesh. Subsequently, we employ Poisson surface reconstruction in Meshlab, instead of Python, for increased efficiency in rendering the mesh and comparing the rendered images with our results.

\noindent\textbf{Look Outside The Room(LOTR)~\cite{Look-outside-the-room}.} Because the LOTR do not generate the novel view images according to the user's input, we use Stable Diffusion to generate the first image and use the first image and the corresponding camera poses to generate the following images. 

\begin{figure*}[t!]
  \centering
  \includegraphics[width=1\linewidth]{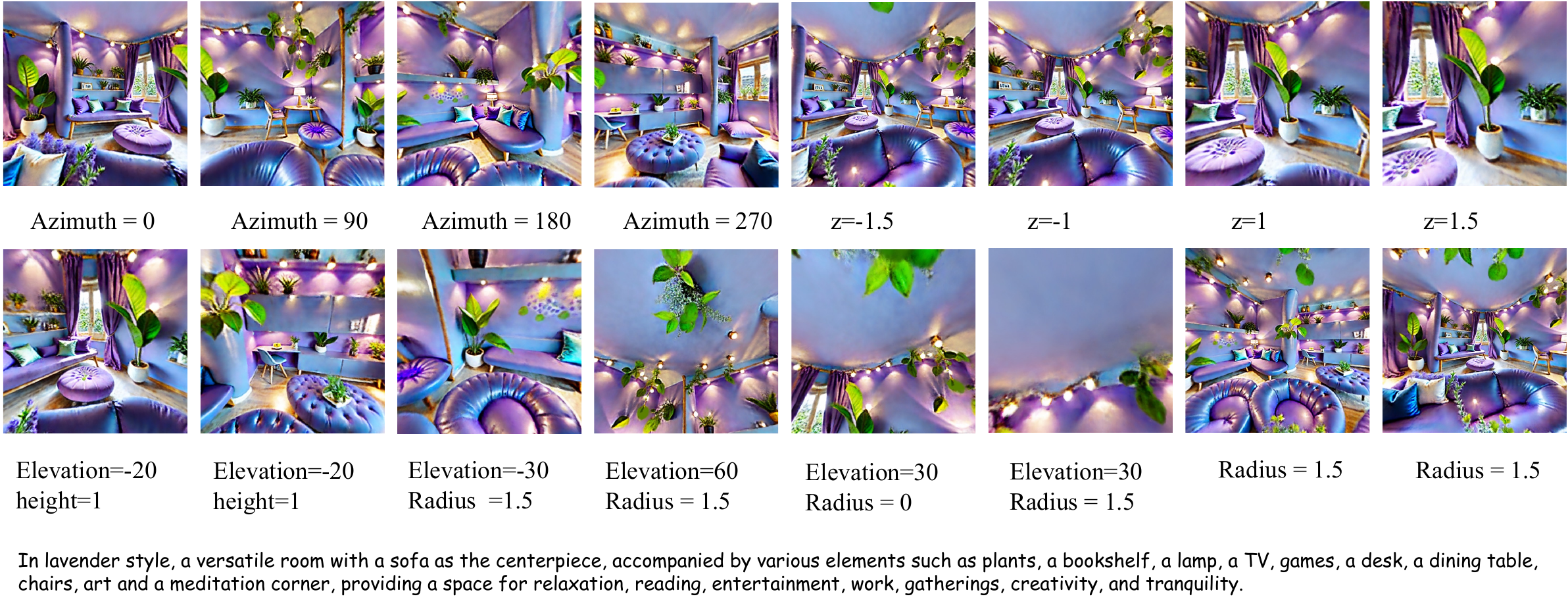}
   \caption{Our more results with 16 views}
   \label{fig:our_result_lavender}
\end{figure*}

\begin{table*}
\vspace{-1mm}
\centering
\resizebox{\linewidth}{!}{

\centering
\setlength{\tabcolsep}{2.5pt}

\begin{tabular}{l||ccccc}

\toprule



& \multicolumn{1}{c}{  DreamFusion }
& \multicolumn{1}{c}{  ProlificDreamer  }
& \multicolumn{1}{c}{  Text2Room }
& \multicolumn{1}{c}{  LOTR  }
& \multicolumn{1}{c}{  ShowRoom3D }

\\
\hline

  Training Time
  &$1\mathrm{h} 30\mathrm{min}$
  &$7\mathrm{h} 20\mathrm{min}$
  &$2\mathrm{h}$
  &$--$
  &$9\mathrm{h}30\mathrm{min}$

  \\ 
  Inference Time
  &$26\mathrm{s}$
  &$31\mathrm{s}$
  &$3 \mathrm{min} 17 \mathrm{s}$
  &$10\mathrm{min}$
  &$27\mathrm{s}$

  \\ \bottomrule

\end{tabular}
}

\caption{
Time comparisons of ShowRoom3D and other state-of-the-art approaches.
\label{tab:time_comparison}
}


\vspace{-5mm}
\end{table*}

\begin{figure}[t!]
  \centering
  \includegraphics[width=1\linewidth]{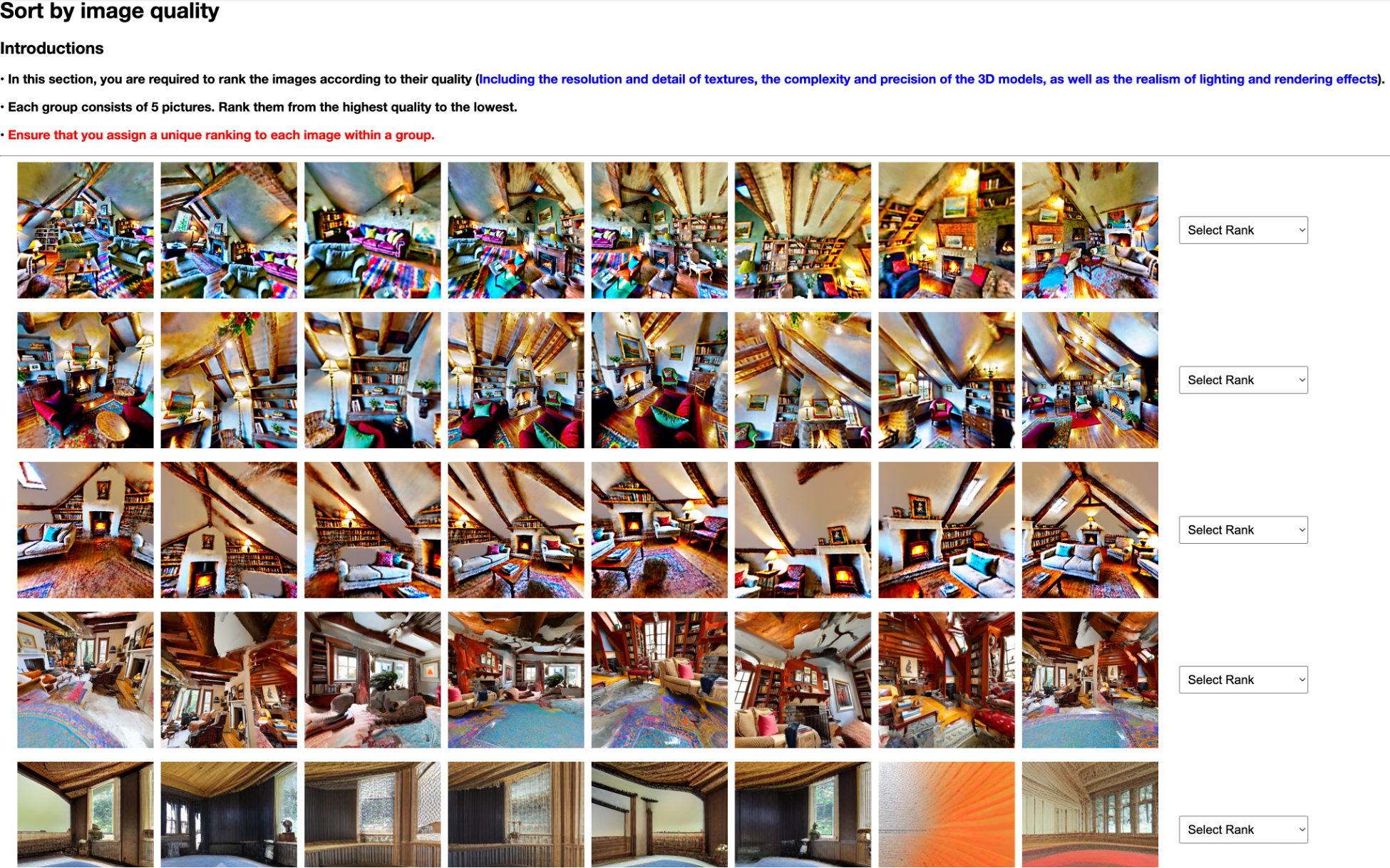}
   \caption{User study interface.}
   \label{fig:user_study}
\end{figure}

\section*{C. More Comparisons with Baselines}
We can see the results in Figure~\ref{fig:baseline_comparison}. We also compare the training time and inference time of our method with other baselines, as depicted in the table.

\noindent\textbf{DreamFusion.} DreamFusion utilizes the vanilla Stable Diffusion~\cite{StableDiffusion} to distill NeRF~\cite{NeRF} in a single stage, leading to several disadvantages. Firstly, the model lacks prior knowledge about 3D indoor scenes, resulting in a severe Janus problem during room generation. As illustrated in Figure~\ref{fig:baseline_comparison}, there are numerous repetitive contents between views, creating a disjointed appearance that does not resemble a cohesive indoor scene but rather a combination of 2D images. Secondly, due to training in a single stage, the room structure is improper, with furniture and the ceiling not consistently placed at the same level.

\noindent\textbf{ProlificDreamer.} ProlificDreamer shares the same disadvantages with DreamFusion. ProlificDreamer incorporates the LORA~\cite{LoRA} model for joint training with NeRF to address oversaturation and enhance content diversity within a single image. However, this approach results in generating more crowded content in one view, exacerbating the Janus problem. The attempt to maintain consistency with the prompt in each image intensifies the overcrowded appearance of the room. Additionally, the increased diversity in every image can lead to more inconsistencies in style between views.

\noindent\textbf{Text2Room.} Text2Room shares the same disadvantages with the aforementioned baselines. Text2Room employs various strategies to fill the generated mesh and ensure its `waterproof' quality, including random camera sampling and Poisson Reconstruction. However, this approach introduces more stretching and blurring artifacts in the rendered images.

\noindent\textbf{Look Outside The Room(LOTR).} LOTR is a novel view synthesis work to generate the next image conditioned on the previous images. However, it has certain disadvantages. Firstly, it struggles to generate images when the rotation degree varies too much, making it ineffective for panoramic image generation. Secondly, LOTR is constrained to generating images in specific directions, and it can not produce corresponding images if the camera is moved backward, up, or down.


\begin{figure*}[t!]
  \centering
  \includegraphics[width=1\linewidth]{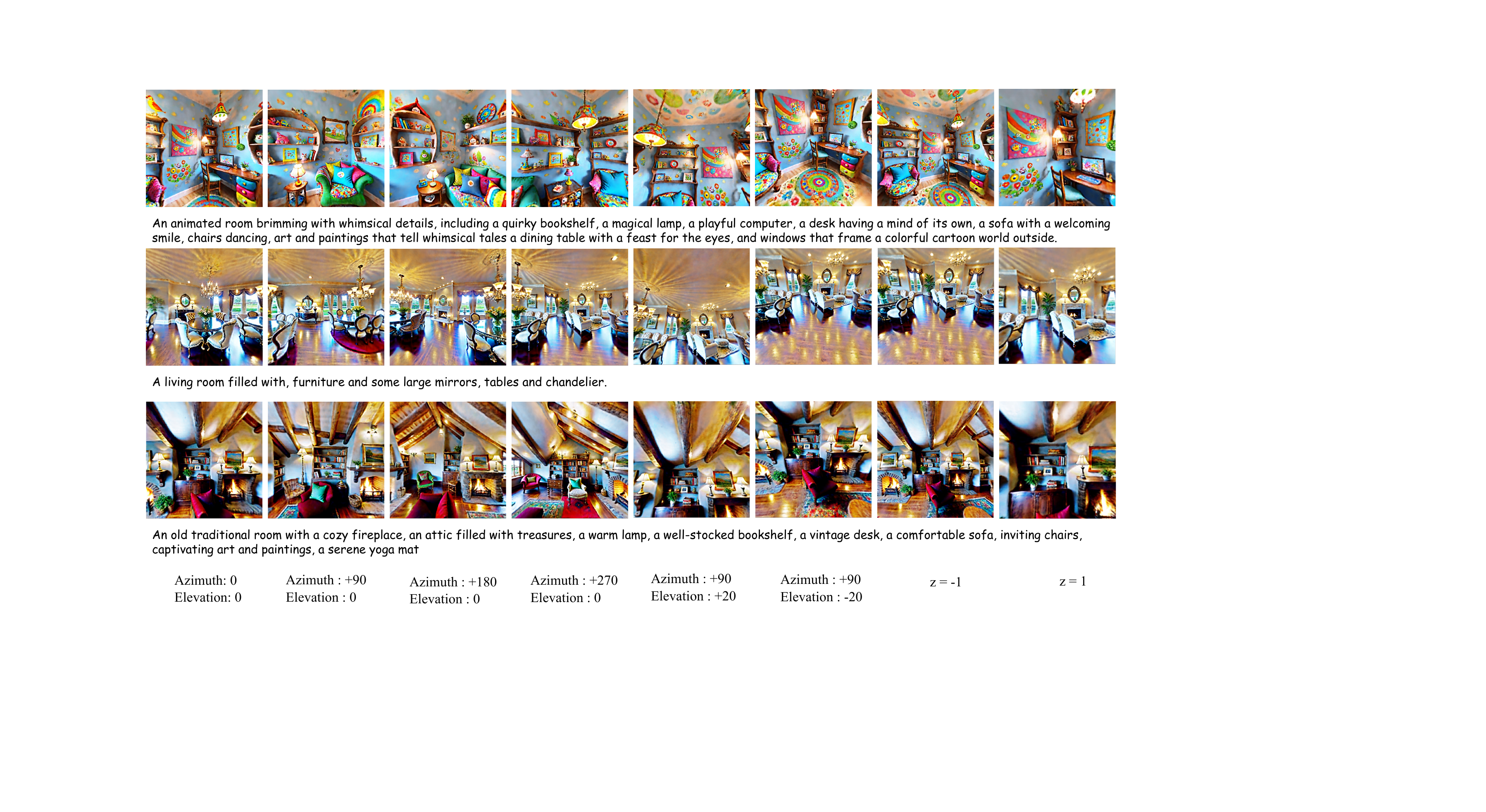}
   \caption{Our more results with 8 views}
   \label{fig:our_result_3scenes}
   \vspace{-3mm}
\end{figure*}

\section*{D. More Results of ShowRoom3D}
In this section, we present an additional set of 16 views for a scene to illustrate that our room can be rendered at any position as shown in Figure~\ref{fig:our_result_lavender}. Furthermore, we provide more results with 8 views to demonstrate that our method is capable of generating diverse types of rooms as shown in Figure~\ref{fig:our_result_3scenes}.

\section*{E. User Study}
We employ our method and other baselines to generate 13 room-scale scenes based on 13 prompts. To ensure fairness, we utilize Amazon MTurk to recruit 24 participants who rank the results on a scale from 5 (highest score) to 1 (lowest score), as depicted in Figure~\ref{fig:user_study}. Users are presented with multiple images from each scene.

\section*{F. Additional Ablation Study}
We show another ablation study about SDS with our training tricks. 

\noindent \textbf{Stable Diffusion In One Stage. } Now we show the results of Stable Diffusion (finetuned on Matterport3D dataset~
\cite{matterport3d}) in one stage. The results will be shown in Figure~\ref{fig:ablation}. In Figure~\ref{fig:ablation}(a), we optimize the NeRF using the Stable Diffusion model (finetuned on the MatterPort3D dataset) in a single stage, incorporating all training tricks. Despite the inclusion of additional training tricks, proper room geometry generation remains elusive. This will demonstrate the effectiveness of our three-stage training pipeline in enhancing the geometry of the room, as shown in Figure~\ref{fig:ablation}(b).

\begin{figure*}[t!]
  \centering
  \vspace{-5mm}
  \includegraphics[width=1\linewidth]{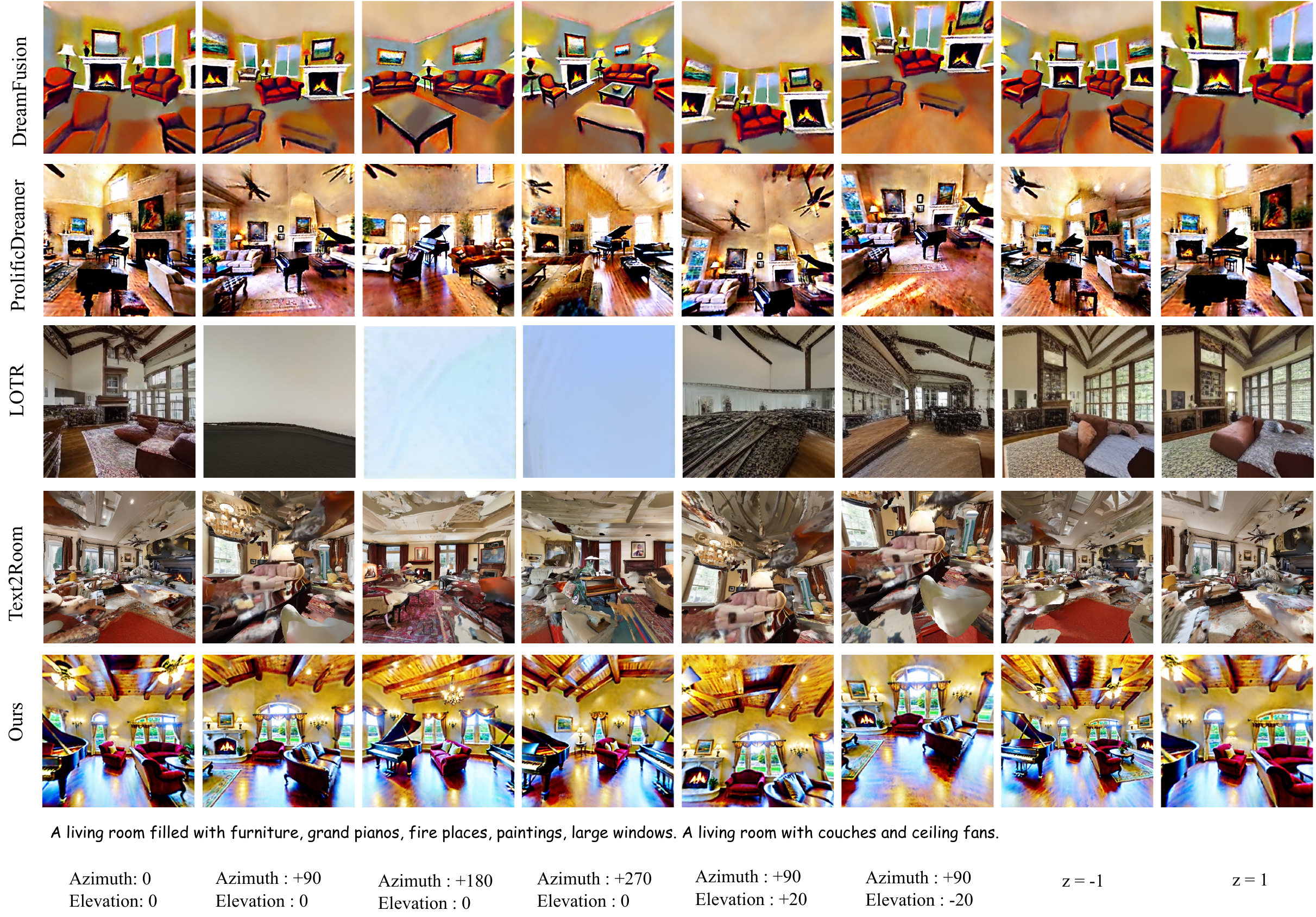}
  \includegraphics[width=1\linewidth]{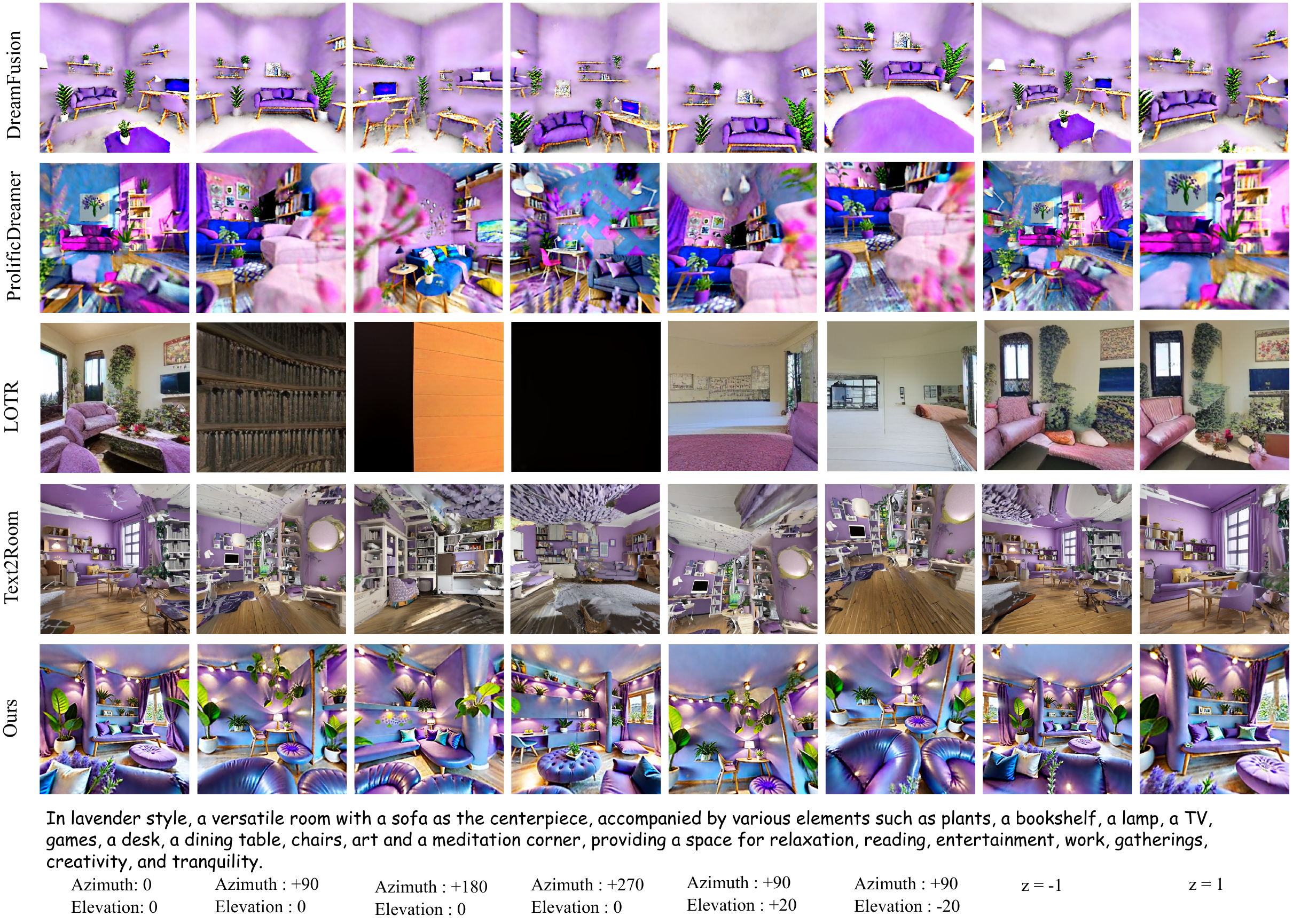}
   \caption{Qualitative comparisons of ShowRoom3D and state-of-the-art approaches.}
   \label{fig:baseline_comparison}
\end{figure*}


\end{document}